\documentclass[a4paper, 11pt, one column, aas_macros]{article}

\usepackage[english]{babel}
\usepackage[utf8x]{inputenc}
\usepackage[T1]{fontenc}
\usepackage{aas_macros}
\usepackage{multicol}

\usepackage[top=0cm, bottom=2.0cm, outer=2.5cm, inner=2.5cm, heightrounded,
marginparwidth=1cm, marginparsep=1cm, margin=1.5cm]{geometry}

\usepackage{caption}
\usepackage{graphicx} %allows you to use jpg or png images. PDF is still recommended
\graphicspath{ {./Images/} }
\usepackage[colorlinks=False]{hyperref} % add links inside PDF files
\usepackage{amsmath}  % Math fonts
\usepackage{amsfonts} %
\usepackage{amssymb}  %
\usepackage{multirow}
\usepackage{subfigure}

\usepackage[square,numbers]{natbib}
\title{Facial Emotion Recognition using CNN in PyTorch}
\author{\small{\rm Deyuan Qu \footnotemark[1]}\\University of North Texas\\DeyuanQu@my.unt.edu 
\and {\rm Sudip Dhakal\footnotemark[2]}\\University of North Texas\\SudipDhakal@my.unt.edu
\and {\rm Dominic Carrillo\footnotemark[3]}\\University of North Texas\\DominicCarrillo@my.unt.edu} 
\date{}
\begin{document}
\maketitle
\footnotetext[1]{Deyuan Qu: Ph.D in Department of Computer Science and Engineering.}
\footnotetext[2]{Sudip Dhakal: Ph.D in Department of Computer Science and Engineering.}
\footnotetext[3]{Dominic Carrillo: Ph.D in Department of Computer Science and Engineering.}

\begin{multicols}{2}
\begin{abstract}
\par In this project, we have implemented a model to recognize real-time facial emotions given the camera images. Current approaches would read all data and input it into their model, which has high space complexity. Our model is based on the Convolutional Neural Network utilizing the PyTorch library. We believe our implementation will significantly improve the space complexity and provide a useful contribution to facial emotion recognition. Our motivation is to understanding clearly about deep learning, particularly in CNNs, and analysis real-life scenarios. Therefore, we tunned the hyper parameter of model such as learning rate, batch size, and number of epochs to meet our needs. In addition, we also used techniques to optimize the networks, such as activation function, dropout and max pooling. Finally, we analyzed the result from two optimizer to observe the relationship between number of epochs and accuracy.
\end{abstract}
\section{Introduction}
\par With the advancement in Machine Learning algorithms and widespread acknowledgment of Computer Vision systems, facial recognition has become an important application. We can leverage the complex features and modules provided by Machine Learning in order to perform improved facial recognition. However, designing the facial recognition system comes with numerous complications. There are multiple application of facial recognition such as business, information security, access control, law enforcement, surveillance system and so on. Application in business include utilization of this type of system for customer satisfaction. For instance, we can monitor the attentive marketing, health status and emotionally intelligent support robotic interface. Such system can facilitate understanding whether customers are satisfied with their services. facial Therefore, researching about facial emotion recognition has increasingly attracted many scientists in computer vision. Human emotions can be recognized through voice, body gesture, or specially facial expression.

\par In 1998, Yann LeCun, a computer scientist working in the field of what is known as Neural Networks, had created a successful application using Convolution Neural Networks (CNNs). Their most known CNN was the LeNet architecture which did character recognition and is able to read zip codes, digits, letters, etc.\cite{LeCun} Since then, CNNs has gone through major development by the computer scientist community from laboratories, academics, entrepreneurs, and business industries e.g. Google and Amazon. An accurate CNNs is sought after within the computer vision topic. This is shown in the amount of different CNNs e.g. AlexNet\cite{AlexNet}, VGG-Net\cite{VGG-Net}, GoogleNet\cite{GoogleNet}, and ResNet\cite{ResNet}. There has been public challenges to create a more accurate model. These challenges were the Facial Emotion Recognition challenge from Kaggle (2013) and the Emotion Recognition in the Wild challenge (2015). From the Kaggle's challenge emerged the FEC2013 dataset, which we will discuss in more detail later, which provides several images of facial emotion expressions. Through-out the years of emotion is universe from six different expression: anger, disgust, fear, happiness, sadness and surprise. However, this universal set is not considered complete thus within current dataset listing you would see seven expressions which added contempt or neutral. FEC2013 utilizes seven expressions and is widely used therefore can be easily compared with other experiments.
\par In this experiment, our goal is to create a CNN to conduct emotion recognition within images and give it a corresponding expressed emotion label. Along with providing us more knowledge and understanding on how a CNN is constructed. From this understanding on CNN, we could provide ideas on the next experiment in where changes could happen to get more deeper into CNN. However, we have to start with getting our basic knowledge.
\par We will discuss related work in Section \ref{SecWork}. The remarkable dataset in Section \ref{SecDataset}. Section \ref{SecMethod} show the methodology of this experiment, in how our CNNs models is constructed in detail. Content in Section \ref{SecResult} is our result. Then close with some conclusion and future work discussion in Section \ref{SecConclusion}.
\section{Related Work} \label{SecWork}
\par As mentioned, major research development is being conducted on facial emotion recognition systems in the past current years. Several approaches have been developed to solve this problem, there has approaches using features-based recognition to deep learning approaches.\cite{reason} However, the CNNs method is widely used for these public challenges and would have high accuracy to detect emotion from the facial expression. From the FER2013 challenge, current leader is Charlie Tang's team, which used several trained CNNs and achieved a 71.162\% accuracy on the test set.\cite{kaggle}
\par In 2016, Bo-Kyeong Kim et al had an test accuracy of 61.6\% and ended up winning the Third Emotion Recognition in the Wild (EmotiW2015) challenge. Their recognition model had a proposed committee of CNNs. There were two strategies from this committee: (1) they would vary the network architecture, the input preprocessing, and the receptive field size in order to obtain more diverse trained deep models, and (2) constructing a hierarchical architecture of the committee which will form a better committee structure and have a better decisional aspect.\cite{emoti} A few other recent examples are from Arriaga et al. in 2017 with a test accuracy of 66\%.\cite{CNNgender} Pramerdorfer and Kampel in 2016 resulted in an accuracy of 75\%, however this accuracy is not listed on the leaderboard on the Facial Expression Recognition Challenge Kaggle's webpage.\cite{CNNart}

\includegraphics[width=.4\textwidth]{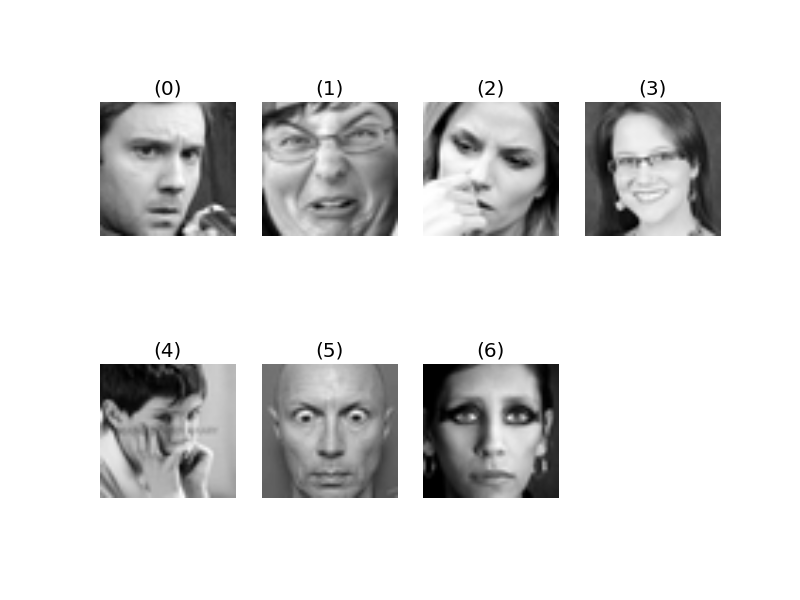}
\captionof{figure}{Example of seven emotions in FER2013 dataset: (0) angry, (1) disgusted, (2) fearful, (3) happy, (4) sad, (5) surprised, (6) neutral}
\label{dataImage}

\includegraphics[width = .5 \textwidth]{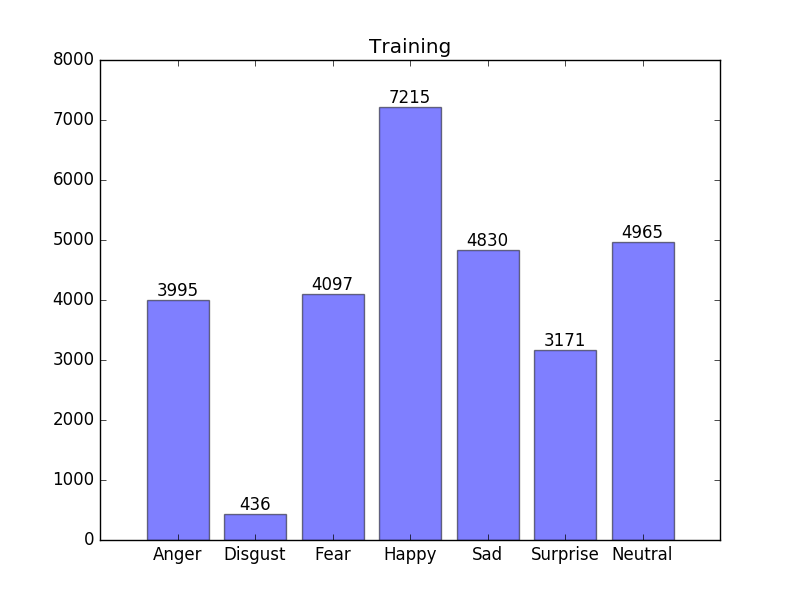}
\captionof{figure}{Overview FER2013 data}
\label{dataPlot}

\section{Dataset} \label{SecDataset}
\par In this paper, we utilizes the CNNs approach for facial emotion recognition. We trained and tested our model on the FER2013 dataset which consists of 48x48 pixel grayscale images of faces. The images are format into a .csv file with 2 columns. The first column is consented of the labels and it has the different 'emotions' (0) angry, (1) disgusted, (2) fearful, (3) happy, (4) sad, (5) surprised, (6) neutral. The second column is the 'data' which are the different pixels in the images. The total count was 30000 images in our dataset.
\par In Figure \ref{dataImage}, we show seven emotional images which are: angry, disgust, fear, happy, sad, surprised, and neutral. Although all the images are from the preprocessing set, we can see that there are various individuals across the entire spectrum of: ethnicity, race, gender and race. These images are also taken at various angles. 
\par The plot in Figure \ref{dataPlot} show the histogram of images of each emotion within our dataset. With disgust being the smallest data (436 images), we assume the prediction of this emotion from the CNNs will be affected in the results of the experiment.
In our experiment we have done data processing following three different methods namely, data separation, data visualization, and data labeling.
\subsection{Data File Formats}
\par We have separated the given dataset into two different .csv files for the convenience of reading and visualizing the dataset for future referencing. The first .csv file consist of all the different labels that we have discussed earlier namely: 'emotions' (0) angry, (1) disgusted, (2) fearful, (3) happy, (4) sad, (5) surprised, (6) neutral. The second .csv file consist of the different pixels or features of the images. 
\subsection{Data Visualization}
\par Once we have separated the dataset, we visualize the features in the image using OpenCV library. We synthesize the 2304 pixel values of each row into each 48*48 emoticon. Our visualization python script will write 30000 emoticons that is equal to the number of images we have in our dataset. The result of the script is the images as shown in the Figure 1. 
\subsection{Data Separation and Labeling}
\par The final part of our data processing includes data separation and labeling. For data separation, we have divided our entire dataset into the training set and the validation set. There are 30000 pictures in total. Among them the first 24000 pictures are under the training set and the remaining are under the validation set. 

\section{Method} \label{SecMethod}
\subsection{Overview}
\par 
The hardware environment of this experiment is as follows: Window10 operating system, Intel i7-7700K CPU, NVIDIA GeForce GTX 1080 Ti GPU hardware environment. The software environment is Python 3.6.8 with PyTorch, and CUDA 11.1 is used to build a convolution neural network. And we used PyCharm for program editing and debugging, PyCharm is one of the most widely used IDEs for Python programming language. In this experiment we used CNN method, because CNNs is the most popular network model among the several deep-learning model available. In addition, CNNs helps us develop researching deep-learning career in the future. And using deep-learning for facial emotion recognition highly reduce the dependence on face-physics-based model and other pre-processing techniques by enabling "end-to-end" learning to occur in the pipeline directly from the input images.

\subsection{CNN}
Traditional convolution neural networks generally include convolution layers, pooling layers and fully connected layers. In the convolution layer, the convolution kernel parameter values are optimized through Backpropagation, and the parameters are shared. Each output of each layer depends on only a small part of the input (sparse connection). Connection can reduce the weight parameters and prevent overfitting from occurring. In the pooling layer, there are maximum pooling and average pooling. Maximum pooling is to select the maximum value of each filter, and average pooling is to select the average value of each filter, after the convolution layer The connection pooling layer can perform feature dimensionality reduction and compress data volume. In the fully connected layer, each neuron is connected to all neurons in the upper layer, and the features extracted by the convolution layer and the pooling layer can be integrated to obtain the probability of each category being accurately identified.

In our experiment we used CNN to recognize seven facial emotion expressions. In the CNN method, the input image is convoluted by the filter set in the convolution layer to generate the feature image. Then, each feature graph is combined into a fully connected network, and the facial expression is recognized as a specific class based output belonging to softmax algorithm.

Our network has three convolution layers and one fully connected layer. Shown in Figure 3. In the first convolution layer, we had 64 filters with kernel size is 3x3. The second convolution is different from first layer. In this layer, we have 128 3x3 filters. The third convolution layers has 256 3x3 filters. These convolution layers also along with batch normalization, max-pooling layer and dropout. Pooling setup is 2x2 with a stride of 1 to reduce the size of the receptive field and avoid overfitting. In dropout layer, a fraction of 0.2 is used.

After these convolution layers, a full connect (FC) layer is added to it after being flattened. FC layer has a hidden layer containing 256 neurons, and the loss function is binary cross entropy (softmax). Similarly, in all layers, the Rectified Linear Unit (ReLU) is used as the activation function to model the nonlinearity. ReLU is easy to use and has excellent performance.

\includegraphics[width = .5 \textwidth]{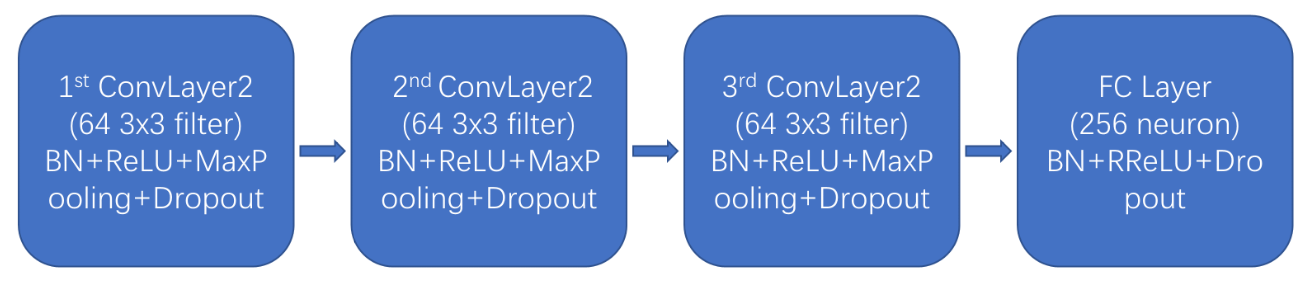}
\captionof{figure}{Architecture of CNN Model}
\label{Deep_CNN}

\section{Result} \label{SecResult}
\subsection{Parameters}
\begin{center}
\captionof{table}{Initial CNN Mode Parameters}
 \begin{tabular}{|c |c| c| c|} 
 \hline
 Batch Size & Epoch & Learning Rate & Decay \\ [0.5ex] 
\hline
 128 & 200 & 0.05 & 1e-5 \\ 
 \hline
\end{tabular}
\end{center}

\par The initialization of model parameters is an important part in the process of neural network training. The appropriate initialization value can make the model converge quickly. The initialization of model parameters in this experiment is shown in Table 1. The batch size is 128, and the total number of training rounds is 200. The optimizer selects the random gradient descent algorithm (SGD), in which the learning rate parameter is set to 0.05, The decay value of learning rate after each update is set to 1e-5.

\includegraphics[width = .5 \textwidth]{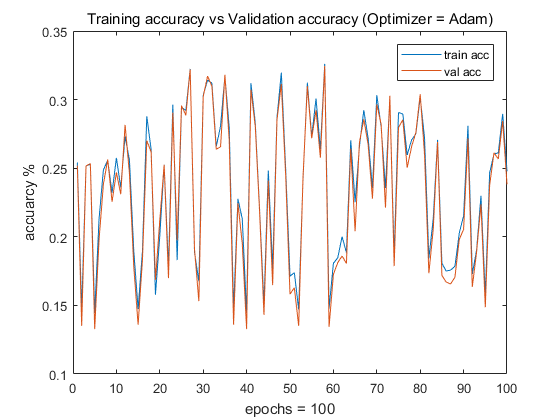}
\captionof{figure}{Tra acc vs Val acc (epochs=100,SGD)}
\label{Data}
\includegraphics[width = .5 \textwidth]{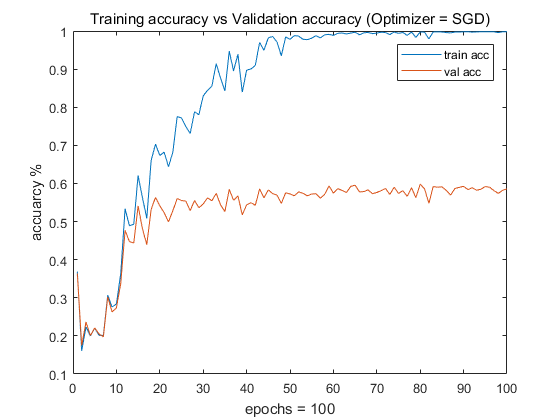}
\captionof{figure}{Tra acc vs Val acc (epochs=200,SGD)}
\label{Data}
\includegraphics[width = .5 \textwidth]{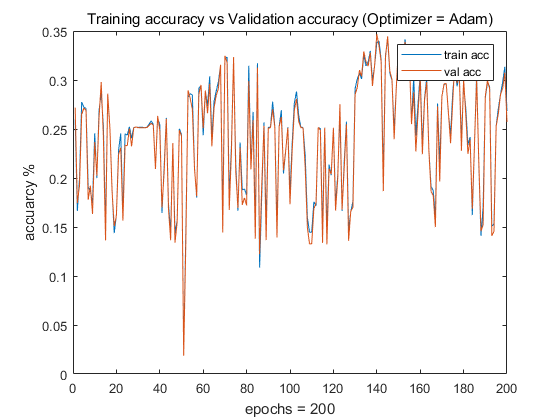}
\captionof{figure}{Tra acc vs Val acc (epochs=100,Adam)}
\label{Data}
\includegraphics[width = .5 \textwidth]{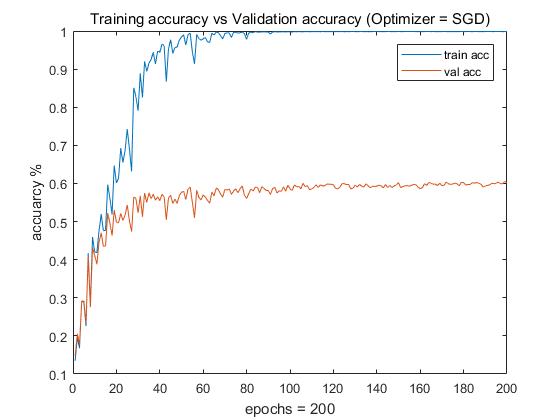}
\captionof{figure}{Tra acc vs Val acc (epochs=200,Adam)}
\label{Data}

In order to verify the feasibility of the proposed scheme, The training dataset and validation dataset are used to adjust the super parameters of the convolution neural network framework for facial expression recognition. When epoch is set to 200, the algorithm can slightly improve the performance of the convolution neural network framework, The change curve of accuracy of training set and verification set is shown in Figure 5. With the increase of epoch times, the accuracy of training set and verification set is improved. Finally, the accuracy of verification set is 60.20\%. The loss value of training set gradually increases with the increase of epoch. When the epoch value is greater than 100, the accuracy of verification set rises slowly, The value of loss almost never decreases. 

\subsection{Analysis}
\par The performance comparison of different parameter settings in this experiment on the Fer2013 data set is shown in Table 2. The first column is the parameters set in the experiment. During the experiment, the parameters such as Batch Size and Learning Rate remain unchanged. When Epoch is set to 100 and the optimizer is SGD, the accuracy of the training set is 99.79\%, the accuracy of the validation set is 58.58\%, and the loss value of the training set is 0.006. In order to improve the training The accuracy on the set and validation set needs to increase the number of training rounds, that means set Epoch to a larger value. When Epoch is set to 200, compared to when Epoch is set to 100, the accuracy of the training set is increased by 0.06\%, and the accuracy of the validation set is increased by 1.62\%.
In order to verify whether other optimizers can improve the accuracy, we use Adam optimizer when other parameters are unchanged. When Epoch is set to 100, the accuracy of the training set is 24.77\%, and the accuracy of the verification set is 23.82\%. When Epoch is set to 200, the accuracy of the training set is 26.83\%, and the accuracy of the validation set is 25.71\%.
It can be seen from the data that when the Adam optimizer is used, no matter whether the epoch is increased or not, the final accuracy rate is very low, and the Loss rate value is also large. It is obvious that using the SGD optimizer is better than the Adam optimizer.

\begin{center}
\captionof{table}{Performance of Different Parameters on Fer2013 Dataset}
 \begin{tabular}{|c|c|c|c|c|} 
 \hline
 Epochs & Optmiz & Acc of tra & Acc of val & LossRate \\ [0.5ex] 
\hline
100 & SGD & 99.79\% & 58.58\% & 0.006 \\ 
 \hline
200 & SGD & 99.85\% & 60.20\% & 0.019 \\ 
 \hline
100 & Adam & 24.77\% & 23.82\% & 1e6 \\ 
 \hline
200 & Adam & 26.83\% & 25.71\% & 4e6 \\ 
 \hline
\end{tabular}
\end{center}

\section{Conclusion} \label{SecConclusion}
In this paper, we have explored CNNs for recognition of facial expressions. Firstly, we implemented a CNN, used the default hyper parameters and got a low accuracy. In order to improve this network and get a higher accuracy, we tuned in different parameters such as the epochs, optimizer and also preprocessed the dataset. The highest accuracy we achieve is 60.20\%. In the future, we will test our CNN model by using other public datasets, in order to get more higher accuracy we will modify both parameters and CNN model. 

\bibliography{refs}
\end{multicols}
\end{document}